\documentclass[11pt, a4paper, logo, onecolumn]{berkeley}

\usepackage[all]{hypcap}
\usepackage{xspace}
\usepackage[capitalize,noabbrev]{cleveref}
\usepackage{subcaption}
\usepackage{wrapfig}
\usepackage{lipsum}
\usepackage{multirow}
\usepackage{tabularx}

\usepackage{url}
\usepackage{float}
\usepackage{tikz}
\usepackage{tabularx} 
\usepackage{array}
\newcolumntype{Y}{>{\centering\arraybackslash}X}
\usepackage{multirow} 
\usepackage{makecell}
\usepackage{subcaption}
\usepackage{hyperref}
\usepackage{amsmath}

\usepackage{graphicx}
\graphicspath{{./images/}}
\usepackage{floatpag} %
\usepackage{graphicx}
\usepackage{amsmath}

\usepackage{etoolbox}

\usepackage[T1]{fontenc}
\usepackage{epsfig} %
\usepackage{amsmath} %
\usepackage{amssymb}  %
\usepackage{wasysym} 
\usepackage{color, soul} %
\usepackage{xcolor}
\usepackage{mathtools}
\usepackage{booktabs}
\usepackage{times}
\usepackage[labelfont=bf, font=footnotesize]{caption}%
\usepackage[labelfont=bf, font=footnotesize]{subcaption}%

\usepackage{lipsum}  
\usepackage{wrapfig}
\usepackage{balance}
\usepackage{algorithm}
\usepackage{setspace}
\usepackage{varwidth}%
\usepackage{textcomp}
\usepackage{tabulary}
\usepackage{relsize}
\usepackage{graphbox} %

\usepackage{import}
\usepackage{tabularx}
\usepackage{array}
\usepackage{makecell}
\usepackage{stackengine}
\usepackage{multirow}
\usepackage{float}
\usepackage{xargs} %
\usepackage[export]{adjustbox}
\usepackage{inconsolata}
\usepackage{url}            %
\usepackage{blindtext}
\usepackage{regexpatch}
\usepackage{bm}
\usepackage{icomma}
\usepackage[title]{appendix}
\usepackage{tikz} %

\definecolor{brightpink}{rgb}{1.0, 0.0, 0.5}
\definecolor{ceruleanblue}{rgb}{0.16, 0.32, 0.75}
\definecolor{coolblack}{rgb}{0.0, 0.18, 0.39}
\definecolor{query}{HTML}{820D07}
\definecolor{meas}{HTML}{2AA63F}
\definecolor{factor}{HTML}{000287}

\definecolor{qcolor}{HTML}{FFA700}
\definecolor{icolor}{HTML}{075493}
\definecolor{pcolor}{HTML}{76D6FF}
\definecolor{fcolor}{HTML}{942193}
\definecolor{prcolor}{HTML}{929292}
\definecolor{lcolor}{RGB}{6,69,173}

\definecolor{contact_color}{HTML}{0B1BE2}
\definecolor{motion_color}{HTML}{FEA700}
\definecolor{shape_color}{HTML}{25A730}
\definecolor{shape_fill_color}{HTML}{D2E1D4}

\usepackage{xspace}

\usepackage{fancyhdr}

\makeatletter
\def\mathcolor#1#{\@mathcolor{#1}}
\def\@mathcolor#1#2#3{%
  \protect\leavevmode
  \begingroup
    \color#1{#2}#3%
  \endgroup
}
\makeatother

\title{OXE-AugE: A Large-Scale Robot Augmentation of OXE 
            for Scaling Cross-Embodiment Policy Learning}
\runningtitle{OXE-AugE: A Large-Scale Robot Augmentation of OXE for Scaling Cross-Embodiment Policy Learning}

\reportnumber{} 

\renewcommand{\today}{Dec 2025} 

\author[1,2]{Guanhua Ji$^*$}
\author[1]{Harsha Polavaram$^*$}
\author[1]{Lawrence Yunliang Chen$^*$}
\author[1]{Sandeep Bajamahal}
\author[1]{Zehan Ma}
\author[1]{Simeon Adebola}
\author[1,3]{Chenfeng Xu}
\author[1]{Ken Goldberg}

\affil[1]{Department of EECS, UC Berkeley}
\affil[2]{GRASP Laboratory, University of Pennsylvania}
\affil[3]{Department of CS, UT Austin}

\correspondingauthor{Lawrence Yunliang Chen (\href{mailto:yunliang.chen@berkeley.edu}{yunliang.chen@berkeley.edu}), Chenfeng Xu (\href{mailto:xuchenfeng@utexas.edu}{xuchenfeng@utexas.edu}).}

\vspace{-20pt}
\makeatletter
\renewcommand{\abscontent}{%
  \noindent
  \begin{center}
    \includegraphics[width=\textwidth,keepaspectratio]{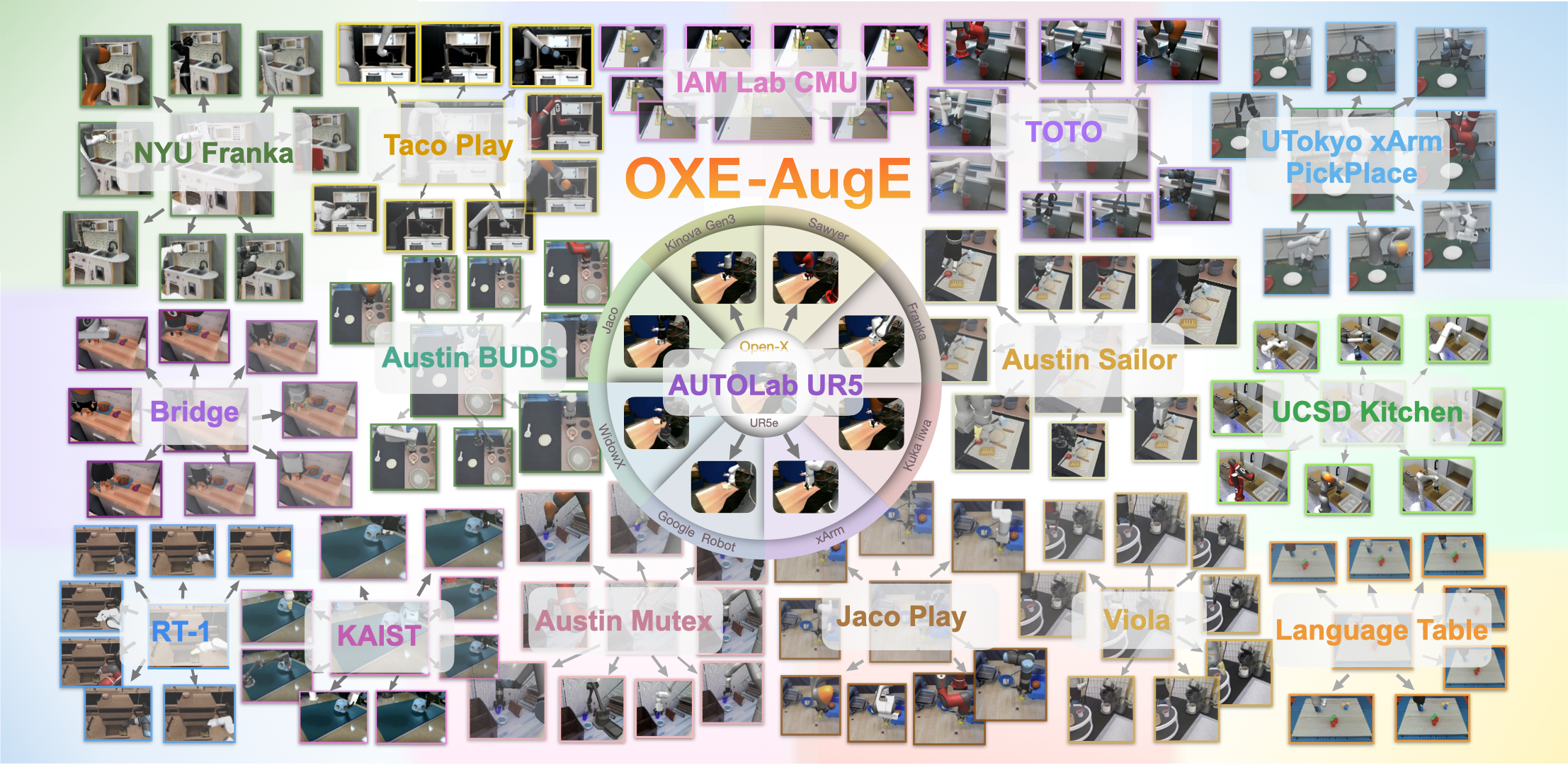}%
    \par\vspace{-0.5em}%
    \captionof{figure}{\textbf{We present OXE-AugE, a large-scale open-source dataset that augments the Open-X Embodiment (OXE) dataset \cite{open_x_embodiment_rt_x_2023} with 9 different robot embodiments across 16 datasets, covering 60\% of the widely-used Octo pretraining mixture \cite{octo_2023}.} In total, OXE-AugE provides over 4 million trajectories, more than triple those in the original OXE. Robots in OXE-AugE include Panda, UR5e, Xarm7, Google robot, widowX, Sawyer, Kinova3, IIWA, and Jaco. We find that training on OXE-AugE improves OpenVLA \cite{kim2024openvla}  and $\pi_0$ \cite{black2024pi_0} policy performance by up to 24-45\% on previously unseen robot-gripper combinations across four real-world manipulation tasks.}
    \label{fig:cover}%
  \end{center}
  \parbox{\dimexpr\linewidth}{\absfont \theabstract}\hfill
  \@ifundefined{@keywords}{}{\vskip1em \noindent \keywordsfont  Keywords: \@keywords}%
}
\makeatother

\begin{abstract}
Large and diverse datasets are needed for training generalist robot policies that have potential to control a variety of robot embodiments---robot arm and gripper combinations---across diverse tasks and environments. As re-collecting demonstrations and retraining for each new hardware platform are prohibitively costly, we show that existing robot data can be augmented for transfer and generalization. The Open X-Embodiment (OXE) dataset, which aggregates demonstrations from over 60 robot datasets, has been widely used as the foundation for training generalist policies. However, 
it is highly imbalanced: the top four robot types account for over 85\% of its real data, which risks overfitting to robot--scene combinations.
We present AugE-Toolkit, a scalable robot augmentation pipeline, and OXE-AugE, a high-quality open-source dataset that augments OXE with 9 different robot embodiments. 
OXE-AugE provides over 4.4 million trajectories, more than triple the size of the original OXE. We conduct a systematic study of how scaling robot augmentation impacts cross-embodiment learning. Results suggest that augmenting datasets with diverse arms and grippers improves policy performance not only on the augmented robots, but also on unseen robots and even the original robots under distribution shifts. In physical experiments, we demonstrate that state-of-the-art generalist policies such as OpenVLA and $\pi_0$ benefit from fine-tuning on OXE-AugE, improving success rates by 24--45\% on previously unseen robot--gripper combinations across four real-world manipulation tasks. Project website: \href{https://OXE-AugE.github.io/}{https://OXE-AugE.github.io/}.
\end{abstract}

\begin{document}

\maketitle



\section{Introduction}
\label{sec:introduction}
Large and diverse datasets have been key to recent progress in general-purpose robot learning, where policies trained on broad experience can generalize to new tasks, objects, and embodiments~\cite{jang2022bc, brohan2023rt1, brohan2023rt2, jiang2022vima, shah2023gnm, shah2023vint, lynch2023interactive, shridhar2021cliport, stone2023open, shridhar2022peract, reed2022a, radosavovic2022real, bharadhwaj2023roboagent, chen2023palix, driess2023palme}. Although performance improves with scale, collecting real-world robot data remains costly and time-consuming~\cite{lee2021beyond, herzog2023deep, kalashnikov2022scaling, jang2022bc, brohan2023rt1, khazatsky2024droid, fang2023rh20t, shafiullah2023dobbe}. The total data volume that can be collected over time is constrained by both the number of teleoperated robots and the duration required to execute and record each trajectory. 
While simulation offers a promising path to scale~\cite{robocasa2024,maddukuri2025simandreal, dai2024automated}, the sim-to-real gap in dynamics and perception of manipulation remains a major challenge~\cite{peng2018sim, ramos2019bayessim, lim2022real2sim2real, memmel2024asid, bousmalis2018using, ho2021retinagan}.

This challenge is amplified as robotic hardware becomes more diverse. New robots with different kinematics, sizes, and grippers are regularly introduced, and policies trained on one platform often fail to transfer to others. Given the high cost of recollecting demonstrations for every new platform, it is desirable to reuse existing data across embodiments. Cross-embodiment generalization---the ability to transfer policies across different robot embodiments---is thus an important goal for scalable and practical robot learning~\cite{yang2023polybot, Doshi24-crossformer, black2024pi_0}.

The Open X-Embodiment (OXE) dataset~\cite{open_x_embodiment_rt_x_2023}, released in 2023, is a major step in this direction. It aggregates demonstrations from over 60 real-world robot datasets collected across different labs, platforms, and tasks. However, most constituent datasets are tied to a single robot in a fixed environment, risking overfitting to robot–scene combinations. Moreover, OXE is highly imbalanced: over 85\% of real trajectories come from just four robots (Franka, xArm, Kuka iiwa, and Google Robot), while many others appear in only 1–2 datasets. Consequently, training a generalist policy on OXE relies on the hope that the policy will implicitly learn embodiment-agnostic features, without explicit mechanisms to mitigate robot bias. In practice, many generalist policies such as Octo~\cite{octo_2023}, OpenVLA~\cite{kim2024openvla}, GR00T~\cite{bjorck2025gr00t}, and $\pi_0$~\cite{black2024pi_0} still require finetuning on new robots, even when they are visually or kinematically similar to those in the training data.

Chen et al.~\cite{chen2024roviaug} propose robot embodiment augmentation---transforming demonstrations collected on one robot into synthetic versions as if performed by another embodiment---using a process called \emph{cross-painting}~\cite{chen2024mirage}. In this work, we improve cross-painting and present AugE-Toolkit, which reduces visual artifacts, improves speed and scalability, while ensuring kinematically valid trajectories using a combination of simulation and learned models.

Using AugE-Toolkit, we generalize robot augmentation beyond pairwise transfer and apply it as a scalable data pipeline. Specifically, we study the effect of scaling robot augmentation and ask: (1) Can robot augmentation improve robustness on the original robot  under visual perturbations?  (2) Does increasing the number of robot augmentations improve performance on augmented robots? (3) Do policies trained on diverse augmentations generalize to unseen robots? 
Results suggest that scaling robot augmentation leads to consistent gains, particularly for generalization to unseen embodiments and visual perturbations. We conjecture that robot augmentation helps policies focus on the spatial geometry between gripper and object, rather than incidental visual features like arm shape or color.

We present \textbf{OXE-AugE}, a high-quality open-source dataset that augments 16 popular OXE datasets with 9 different robot embodiments. The resulting dataset provides over 4 million trajectories—more than triple the size of the original OXE—and covers 60\% of the widely used Octo pretraining mixture. By varying the robot embodiment while preserving task and scene, OXE-AugE provides a new resource for training robust and transferable visuomotor policies.

We evaluate whether state-of-the-art generalist models such as $\pi_0$~\cite{black2024pi_0} and OpenVLA-OFT~\cite{kim2025fine} can benefit from OXE-AugE. We fine-tune each model on the augmented dataset and evaluate them on a real Franka robot equipped with two distinct grippers. Across four manipulation tasks, we observe 24--45\% improvements in success on previously unseen robot–gripper configurations, suggesting the practical utility of large-scale robot augmentation.

This paper makes 4 contributions:
\begin{enumerate}
    \item AugE-Toolkit, an improved and easy-to-use robot augmentation pipeline that enables scalable, high-quality augmentation.
    \item OXE-AugE, a large open-source dataset that augments OXE with 9$\times$ more robot embodiments across 16 datasets, totaling over 4 million trajectories and covering 60\% of the Octo pretraining mixture.
    \item A simulation study of how scaling robot augmentation affects generalization to both seen and unseen embodiments, and robustness to visual perturbations.
    \item Physical experiments suggesting that fine-tuning foundation models on OXE-AugE can improve zero-shot success by 24--45\% on novel robot embodiments.
\end{enumerate}

\section{Related Work}

\subsection{Cross-Embodiment Robot Learning}
A core challenge in generalist robot learning is how to generalize across robot embodiments without collecting new data for each platform.
One common approach is domain randomization, where physical parameters of the robot (e.g., joint and link properties) are randomized in simulation to learn robot-conditioned policies~\cite{yu2023multi, chen2018hardware, shao2020unigrasp, xu2021adagrasp, wang2018nervenet, sanchez2018graph, pathak2019learning, huang2020one, kurin2020my}.
Hu et al.~\cite{hu2021know} learn a world model with the robots masked out, and use visual MPC during execution time when deployed on a new robot.
Another line of work explores using human data. Recent efforts have leveraged human videos~\cite{xiong2021learning, bahl2022human, duan2023ar2, lepert2025phantomtrainingrobotsrobots, kareer2024egomimicscalingimitationlearning} for robot manipulation, and motion retargeting methods~\cite{he2025asap, liao2025beyondmimic, allshire2025videomimic, yin2025visualmimic, yang2025omniretarget} have been applied in locomotion settings.
Other work has also explored pooling large and diverse data, including from different robots~\cite{kalashnikov2018qt, levine2018learning, dasari2019robonet, acronym2020, fang2023rh20t, walke2023bridgedata, bousmalis2023robocat} and found that the resulting policies are generalizable to new tasks and embodiments~\cite{alayrac2022flamingo, jang2022bc, stone2023open, jiang2022vima, reed2022a, radosavovic2022real, shah2023gnm, yang2023polybot, bharadhwaj2023roboagent, brohan2023rt1, brohan2023rt2, chen2023palix, driess2023palme}. 

The Open X-Embodiment project~\cite{open_x_embodiment_rt_x_2023}, in particular, aggregated more than 60 datasets and demonstrated the benefit of training on various embodiments through experiments in multiple labs. Many have leveraged the OXE dataset and developed ``generalist policies'' that can perform multiple tasks on a wide range of robots~\cite{octo_2023, kim2024openvla, black2024pi_0, yang2024pushing, black2025pi05, bjorck2025gr00t, wen2025diffusionvla}. However, the robot types in OXE are severely unbalanced, and trained policies typically perform much better on robots that are well-represented in the datasets and still require a fair amount of finetuning data to transfer to a new robot. Mirage~\cite{chen2024mirage} proposes a test-time image inpainting pipeline to replace the new target robot in the image with the familiar source robot seen during training to achieve zero-shot cross-embodiment transfer. In this work, we inpaint the opposite direction and improve the pipeline for scalable data augmentation.

\subsection{Augmenting Real Robot Data}
Given the high cost of collecting real robot data, many approaches have explored augmenting existing datasets. 
Real-to-sim-to-real pipelines~\cite{lim2022real2sim2real, torne2024reconciling, torne2024robot, li2024robogsim, ye2025video2policy, pfaff2025scalable, dan2025xsim, geng2025roboverse, zhao2025robot, zhang2025robowheel} reconstruct 3D object meshes from real videos and tune simulation parameters to build digital twins or ``digital cousins''~\cite{dai2024automated} for policy learning. Some works~\cite{yu2025real2render2real, yang2025novel} avoid simulating the physics but still require an extra step of taking multi-view images of the scene or objects to build 3D Gaussian Splatting (3DGS). Many works that leverage 3DGS for rerendering and trajectory synthesis~\cite{zhou2301nerf, pan20251001demos, zhang2024diffusion} are most applicable to eye-in-hand images. 

Another category of work applies 2D image or video generation models to augment directly in image space. Examples include editing backgrounds or objects~\cite{yu2023scaling, chen2023genaug, mandi2022cacti, bharadhwaj2023roboagent}, or synthesizing novel views~\cite{chen2024roviaug, tian2024view}. For robot transfer, Shadow~\cite{lepert2024shadow} masks out robots in images; RoVi-Aug~\cite{chen2024roviaug} applies diffusion models to transform one robot appearance into another; and some works~\cite{lepert2025phantomtrainingrobotsrobots, lepert2025masquerade, li2025h2r, yang2025x, ci2025h2r} transform human videos into robot videos. However, most of these works focus on one-to-one transfer between a known source and target embodiment. In contrast, we study \emph{scaling} robot augmentation across many target embodiments and investigate how such augmentation impacts generalization and robustness in both simulation and real-world settings.

\subsection{Study of Scaling in Robot Learning}

Several recent studies have analyzed how performance in robot learning scales with data volume and model capacity~\cite{sartor2024neural}. Models like VIMA~\cite{jiang2022vima}, RT-1~\cite{brohan2023rt1}, Octo~\cite{octo_2023}, and HPT~\cite{wang2024scaling} report consistent gains when increasing dataset size and model scale. For data scaling, Lin et al.~\cite{lin2024data} show approximate power-law improvements in generalization as the number of environments and objects increases, with diversity often more valuable than additional demonstrations per setting. Saxena et al.~\cite{saxena2024matters} further dissect scaling effects by quantifying contributions from camera viewpoint, spatial layout, and object variety.
For robot embodiment scaling, OXE~\cite{open_x_embodiment_rt_x_2023}, RoboCat~\cite{bousmalis2023robocat}, and CrossFormer~\cite{Doshi24-crossformer} find that training on multiple robot types leads to better transfer than training on a single target robot alone. Yang et al.~\cite{yang2024pushing} show benefits from jointly training across manipulation and navigation domains.
In this work, rather than pooling additional real data from multiple robots, we study the effect of scaling \emph{robot augmentation}—synthetically generating data from multiple robot embodiments—on performance across original, augmented, and entirely unseen robot configurations.

\section{Problem Statement}\label{sec:problem_statement}

We consider the standard imitation learning setting~\cite{pomerleau1989alvinn, ross2011reduction}, where we have a demonstration dataset $\mathcal{D}^{\mathcal{S}} = \{\tau_1^{\mathcal{S}}, \tau_2^{\mathcal{S}}, ..., \tau_n^{\mathcal{S}}\}$ consisting of $n$ successful trajectories performed by a source robot $\mathcal{S}$. Each trajectory $\tau_i^{\mathcal{S}} = (o_{1:H_i}^{\mathcal{S}}, p_{1:H_i}^{\mathcal{S}}, a_{1:H_i}^{\mathcal{S}})$ contains RGB observations $o_t^{\mathcal{S}}$, corresponding gripper 6D poses $p_t^{\mathcal{S}}$, and actions $a_t^{\mathcal{S}}$ for timesteps $t = 1, \dots, H_i$.

We study \textit{robot augmentation}—synthetically transforming each trajectory $\tau_i^{\mathcal{S}}$ into a corresponding trajectory $\tau_i^{\mathcal{R}}$ for a different robot embodiment $\mathcal{R}$ (arm and gripper) performing the same task. Given $\mathcal{D}^{\mathcal{S}}$ and the kinematic models of the robots (e.g., URDF), this yields a synthetic dataset $\mathcal{D}^{\mathcal{R}}$ with aligned images, poses, and actions $(o_{1:H_i}^{\mathcal{R}}, p_{1:H_i}^{\mathcal{R}}, a_{1:H_i}^{\mathcal{R}})$. Following prior work~\cite{chen2024roviaug}, we assume the grippers across robots are similar in shape and function (e.g., 2- or 3-jaw grippers), enabling all robots to perform the same task with a shared strategy. Similar to prior work~\cite{shah2023gnm, chen2024mirage, yang2023polybot, yang2024pushing}, we use Cartesian control and assume that the coordinate frames of all robots are known, allowing alignment through rigid transformations.

Let $\mathcal{R}_\text{Aug} = \{\mathcal{R}_1, \mathcal{R}_2, \dots, \mathcal{R}_N\}$ denote the set of robot embodiments used for augmentation. We train a policy on the resulting augmented dataset $\mathcal{D}^{\text{Aug}} =  = \bigcup_{i=1}^{N} \mathcal{D}^{R_i}$ and evaluate it on a target robot $\mathcal{T}$ at test time. We examine how robot augmentation affects: (1) \textbf{Robustness:} $\mathcal{T} = \mathcal{S}$ (original source robot), (2) \textbf{Transfer:} $\mathcal{T} \in \mathcal{R}_\text{Aug}$ (performance on augmented robots), and (3) \textbf{Generalization:} $\mathcal{T} \notin \mathcal{R}_\text{Aug}$ (unseen robots).
We further study how these outcomes vary as we scale the number of augmented embodiments, from a single target robot ($|\mathcal{R}_\text{Aug}| = 1$) to $N$ distinct robot types.

\begin{figure*}[t!]
    \centering
    \includegraphics[width=\textwidth,keepaspectratio]{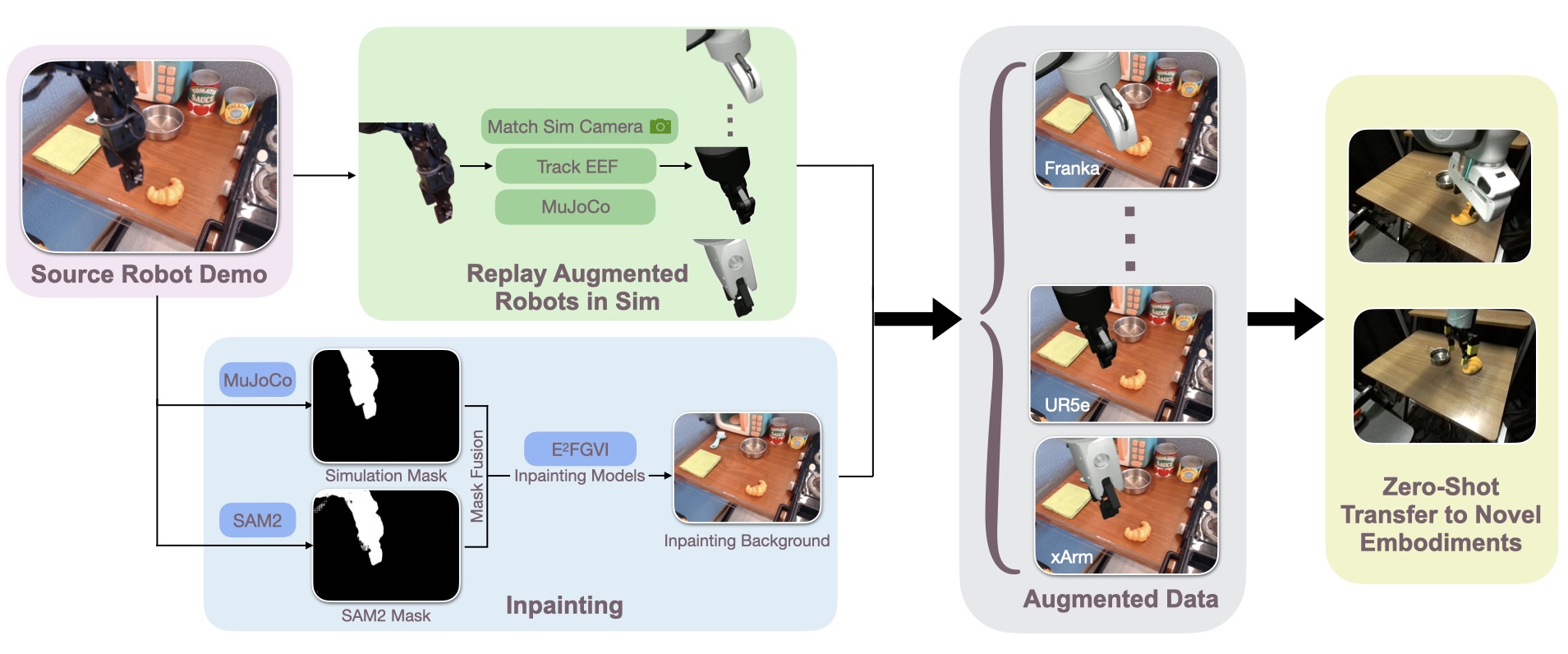}
    \caption{\textbf{AugE-Toolkit pipeline.} Given a source robot image and its corresponding robot poes, AugE-Toolkit fuses a learned SAM2 mask with a simulation-rendered mask to segment the robot, inpaints the background via E$^2$FGVI~\cite{liCvpr22vInpainting}, and replays the same trajectory with another robot in simulation~\cite{zakka2025mujoco}. The augmented robot is composited into the reconstructed scene to form the augmented video. Policies are trained on both real and augmented data and evaluated on unseen embodiments.}
    \label{fig:pipeline}
\end{figure*}

\section{Methods}
\label{sec:methods}

\subsection{Preliminaries: Cross-Painting Framework}

Cross-painting~\cite{chen2024mirage, chen2024roviaug, lepert2025phantomtrainingrobotsrobots} is a three-stage pipeline applied to each image in a robot trajectory:  
(i) \emph{source-robot segmentation}, (ii) \emph{background inpainting}, and (iii) \emph{augmented-robot replay and compositing}.  
The goal is to replace the robot embodiment in each frame while preserving task and scene context.

Existing implementations differ mainly in how each stage is realized.  
Learning-based methods use diffusion models to modify the robot directly in pixel space~\cite{chen2024roviaug}, requiring no explicit calibration and producing visually realistic results. However, they lack kinematic guarantees and scale poorly to many target robots, since each new embodiment typically requires a separately trained model.  
Simulation-based methods~\cite{chen2024mirage} render the robot using known camera parameters, ensuring geometric fidelity. These methods were originally designed for test-time adaptation, where accurate calibration is feasible, but are difficult to apply to large-scale offline datasets that often lack such information.

We present AugE-Toolkit, which builds on this framework to enable scalable robot augmentation.  
It retains the physical accuracy of simulation-based rendering but is also applicable to uncalibrated or coarsely aligned data.  
Through mask fusion and semi-automatic base tuning, AugE-Toolkit synthesizes physically consistent robot augmentations across many embodiments with minimal manual effort.

\subsection{AugE-Toolkit: Scalable Robot Augmentation}

AugE-Toolkit (Fig.~\ref{fig:pipeline}) extends cross-painting in RoVi-Aug~\cite{chen2024roviaug} with three key components.

\noindent \textbf{(1) Fusion of Simulation and Learned Masks.}  
To obtain accurate robot masks without camera calibration, we combine simulation-based and learned segmentation. We fine-tune SAM2~\cite{ravi2024sam2} on a small labeled subset (20 trajectories from each of 16 OXE datasets).  
While the learned masks align well with image appearance, they often over- or under-segment near the gripper.  
Simulation masks are geometrically accurate but may be globally misaligned due to unknown camera poses.  
We align and fuse them in three steps:  
(1) \emph{Translation alignment}: shift the simulation mask within a small grid to maximize IoU with the learned mask;  
(2) \emph{Distance pruning}: remove learned-mask pixels farther than a threshold $\tau$ from the aligned simulation boundary;  
(3) \emph{Union and smoothing}: combine both masks and apply morphological closing.  
This fusion process corrects for calibration errors and enables accurate rendering even on uncalibrated datasets. Downstream training can also flag and filter data whose final masks' IoU are large.

\noindent \textbf{(2) Automatic Base Position Tuning.}  
To accommodate robots with different kinematic reach and scale, we automatically adjust the base position of each target robot in simulation to ensure all end-effector poses from the source trajectory are reachable.  
Starting from an initial base pose, we iteratively sample offsets $\pm\Delta$ along the $(x, y, z)$ axes, compute tracking error, and halve the step size until the maximum error falls below 1\,cm or a maximum iteration count is reached.  
Trajectories without feasible base positions are discarded.\footnote{In practice, two datasets (RT-1 Fractal and Language Table) have many trajectories that are unreachable by the WidowX robot; we exclude these from their augmentations. All other datasets are augmented into all 9 robot-gripper combinations, with >95\% of trajectories achieving replay errors under 0.25\,cm. See Appendix for details.}  
This procedure enables consistent augmentation across compact arms (e.g., WidowX) and larger robots (e.g., Google Robot) while preserving motion fidelity.\footnote{For mobile robots, the optimized base translation can be interpreted as a movement action.}

\noindent \textbf{(3) Scalable Multi-Robot Deployment.}  
We employ simulation rendering rather than generative synthesis to ensure temporal coherence and physical validity.  
Generative models often introduce flicker and geometric artifacts, whereas URDF-based rendering guarantees pose accuracy and realism.  
Our pipeline is implemented on MuJoCo Playground~\cite{zakka2025mujoco}, which supports a large collection of robots~\cite{menagerie2022github}. Adding a new robot only requires registering its model; no retraining or calibration is needed.

Each stage of the pipeline is embarrassingly parallel. Since each target robot independently replays the same trajectory, augmentation across multiple robots can be performed concurrently.  
A 50-frame $640 \times 480$ clip completes in approximately 25 seconds per robot; with 32-way parallelism, throughput reaches up to 75 clips per minute.

\begin{figure*}[t]
    \centering
    \includegraphics[width=\textwidth,keepaspectratio]{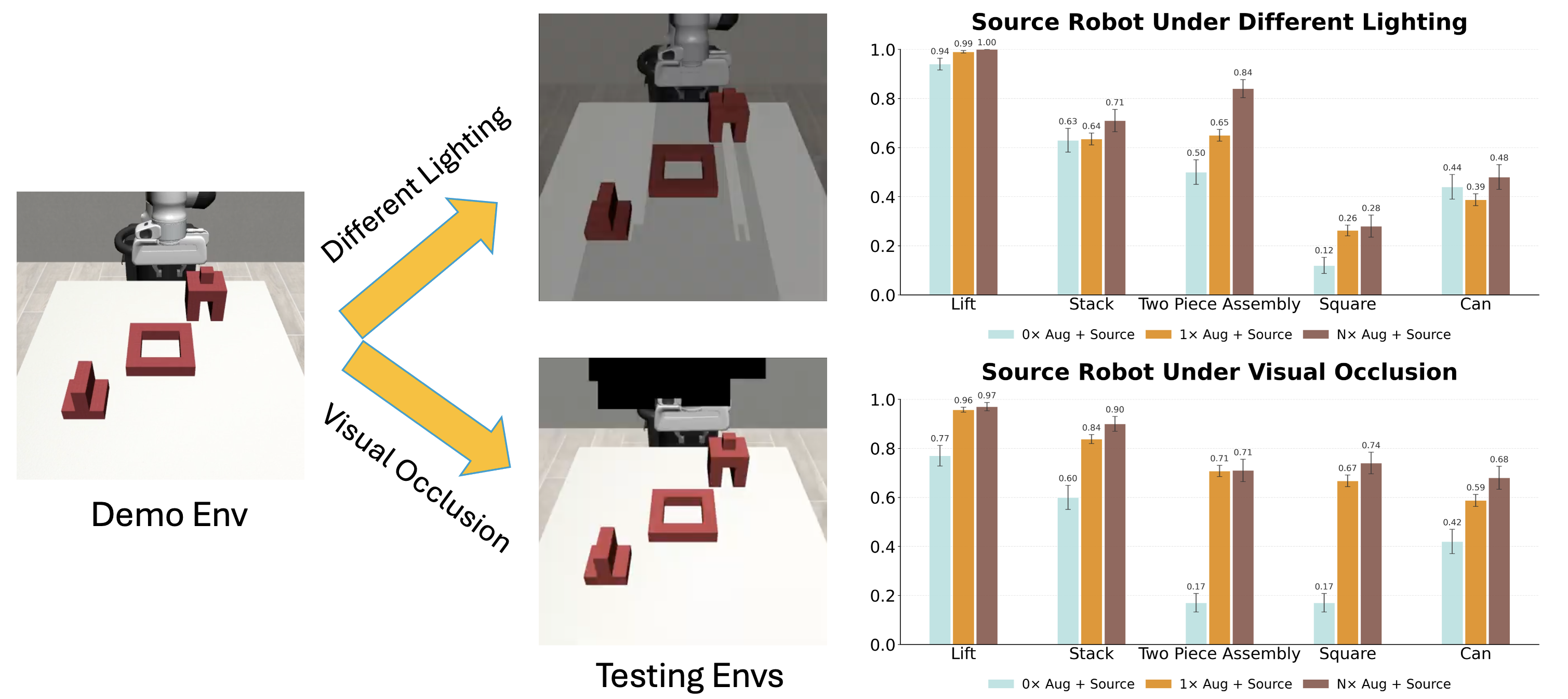}
    \caption{\textbf{Robot augmentation improves robustness on the source robot under visual perturbations.} \textbf{Left:} We consider two types of perturbations: different lighting conditions and visual occlusions. \textbf{Right: } Performance of policies trained on various augmented datasets on the Franka (source) robot. The performance of policies without augmentation severely degrades, while increasing the number of augmented robots makes policies more robust.}
    \label{fig:sim_result_source_robot}
\end{figure*}

\begin{figure*}[t]
    \centering
    \includegraphics[width=\textwidth,keepaspectratio]{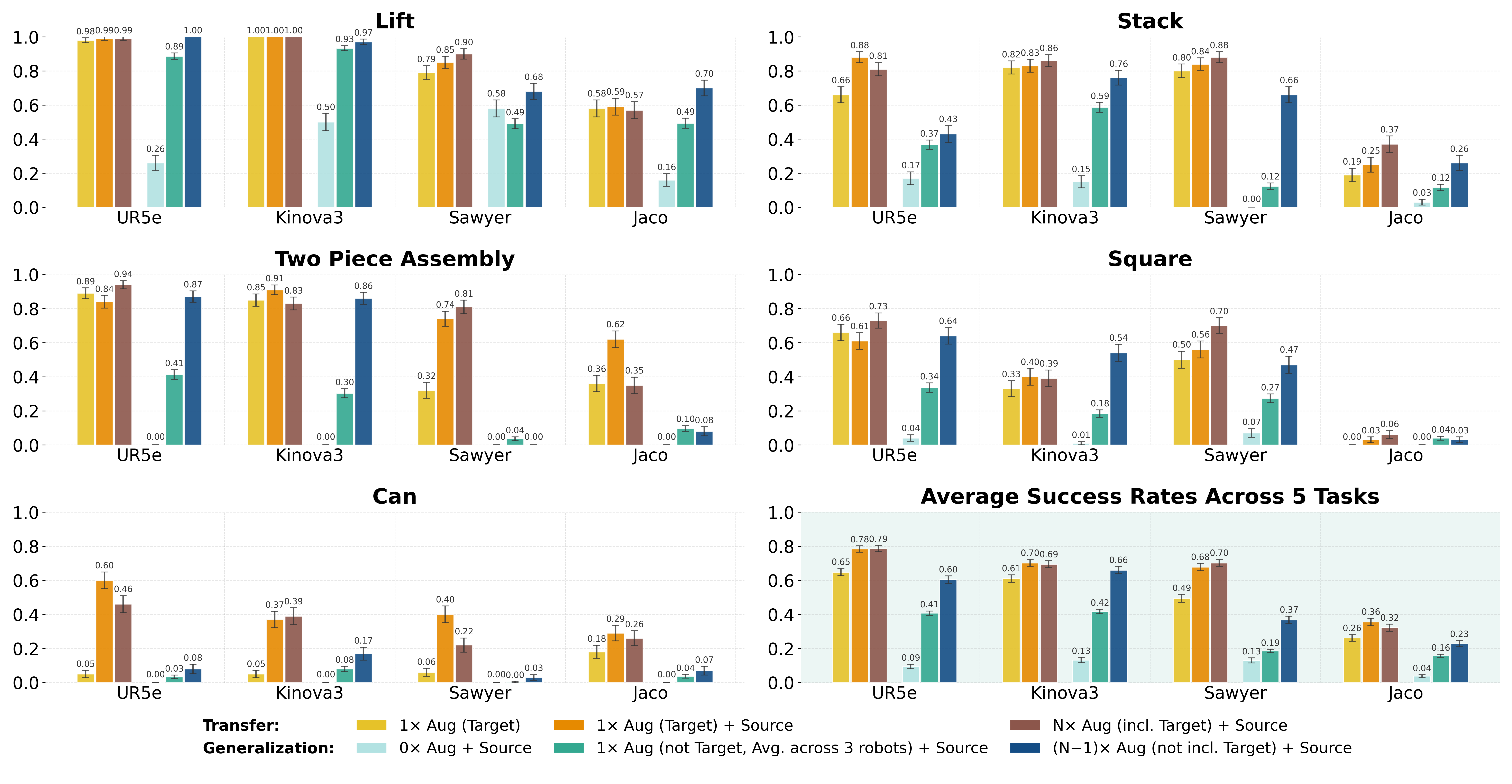}
    \caption{\textbf{Simulation experiments on scaling robot augmentation.} 
    We evaluate how scaling the number of augmented robots impacts (1) \textbf{Transfer}: performance on augmented robots (orange), and (2) \textbf{Generalization}: performance on unseen robots not included in the augmented set (blue). For transfer, we compare policies trained only on the augmented target robot (``1× Aug (Target)''), on both the source data and that target robot (``1× Aug (Target) + Source''), and on the source data plus all $N$ augmented robots (``$N$× Aug (incl. Target) + Source''). For generalization, we compare policies trained only on the source data (``0× Aug + Source''), on the source data and one non-target augmented robot (``1× Aug (not Target) + Source''), and on the source data plus $N\!-\!1$ augmented robots, leaving the target out (``($N$-1)× Aug (not incl. Target) + Source''). 
    In the transfer setting, ``1× Aug + Source'' substantially outperforms ``1× Aug,'' and ``$N$× Aug + Source'' achieves comparable or slightly better performance. In the generalization setting, performance improves consistently with augmentation diversity, with ``($N$-1)× Aug (not incl. Target) + Source'' often rivaling ``1× Aug (Target).''}
    \label{fig:sim_main_results}
\end{figure*}

\section{Scaling Robot Augmentation: A Systematic Study in Simulation}
\label{sec:sim_study}

We begin with a systematic simulation study to examine how robot augmentation scales in terms of transfer, generalization, and robustness. While prior work~\cite{chen2024roviaug} has shown that augmenting demonstrations from a source robot to a known target enables zero-shot transfer, our goal is to investigate whether robot augmentation provides broader benefits when scaled across multiple target robots.

We follow the Mirage evaluation setup~\cite{chen2024mirage} and consider five Robosuite~\cite{robosuite2020} tasks: \textsc{Lift}, \textsc{Stack}, \textsc{Can}, \textsc{TwoPiece Assembly}, and \textsc{Square}. For each task, we use 200 demonstrations from Robomimic~\cite{robomimic2021} (\textsc{Lift}, \textsc{Can}, \textsc{Square}) or MimicGen~\cite{mandlekar2023mimicgen} (\textsc{Stack}, \textsc{TwoPiece Assembly}), all collected on a Franka robot (source $\mathcal{S}$ = Franka). Using our pipeline, we augment the demonstrations into four additional robot embodiments: UR5e, Kinova Gen3, Sawyer, and Jaco (which has a 3-jaw gripper). We use Diffusion Policy~\cite{chi2023diffusion} to train separate policies for each condition using RoboMimic~\cite{robomimic2021}; see Appendix for training details.

\paragraph{Training configurations.}
We consider four data regimes: 
\begin{itemize}
    \item \textbf{No Augmentation (``0× Aug + Source''):} Train only on the original Franka demonstrations ($\mathcal{D}^{\text{Train}} = \mathcal{D}^{\mathcal{S}}$).
    \item \textbf{1 Robot Augmentation (``1× Aug''):} Augment the source data into a single target robot $\mathcal{R}$ and train only on the augmented demonstrations ($\mathcal{D}^{\text{Train}} = \mathcal{D}^{\mathcal{R}}$). This is the standard setting studied in prior work. 
    \item \textbf{1 Robot Augmentation Together with Source Data (``1× Aug + Source''):} Combine source and one target robot’s data ($\mathcal{D}^{\text{Train}} = \mathcal{D}^{\mathcal{S}} \cup \mathcal{D}^{\mathcal{R}}$).
    \item \textbf{Multi-Robot (``N× Aug + Source''):} Combine source data with all $N$ augmented robots ($\mathcal{D}^{\text{Train}} = \mathcal{D}^{\mathcal{S}} \cup \bigcup_{i=1}^{N}\mathcal{D}^{\mathcal{R}i}$, $\mathcal{R}_i \in  \{$UR5e, Kinova Gen3, Sawyer, Jaco$\}$).
\end{itemize}

\paragraph{Evaluation protocols.}
Following Sec.~\ref{sec:problem_statement}, we evaluate each policy under three conditions:
\begin{itemize}
    \item \textbf{Robustness (Source robot):}  
    Evaluate on the original source robot ($\mathcal{T} = \mathcal{S}$) under visual perturbations (lighting shifts and occlusions).  
    We compare policies trained on ``0× Aug + Source,'' ``1× Aug + Source,'' (averaged across 4 augmented robots) and ``$N$× Aug + Source.”

    \item \textbf{Transfer (Augmented robots):}
    Evaluate on robots used for augmentation ($\mathcal{T} \in \mathcal{R}_\text{Aug}$).
     We compare ``1× Aug (Target),'' ``1× Aug (Target) + Source,'' and ``$N$× Aug (incl. Target) + Source'' settings.
    
    \item \textbf{Generalization (Unseen robots):}  
    Evaluate on robots not used for augmentation ($\mathcal{T} \notin \mathcal{R}_\text{Aug}$).  
    We compare ``0× Aug + Source,'' ``1× Aug (not Target) + Source,'' (averaged across 3 augmented robots) and ``($N$-1)× Aug (not incl. Target) + Source,'' where ``$N$–1'' excludes one hold-out robot from training and evaluates on it.

\end{itemize}

\begin{figure*}[t]
    \centering
    \includegraphics[width=\textwidth,keepaspectratio]{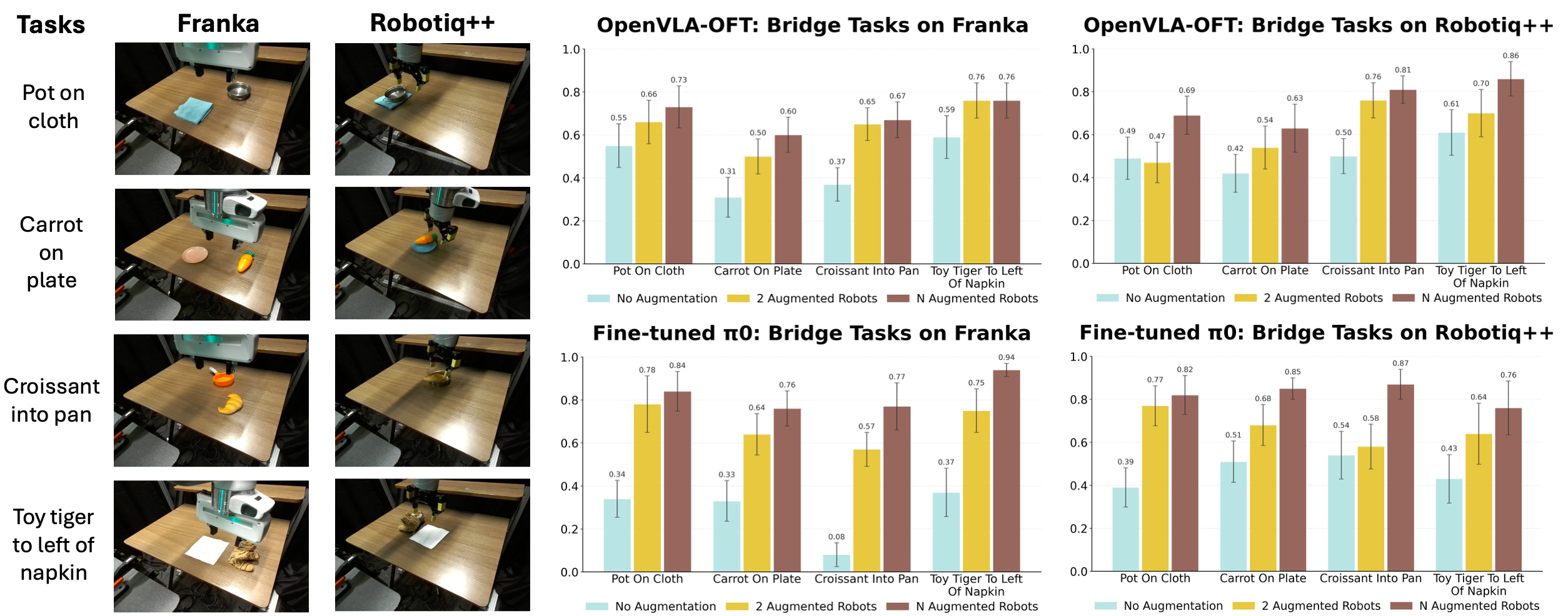}
    \caption{\textbf{Physical Experiments.} \textbf{Left}: Illustration of the 4 tasks and the 2 testing embodiments. ``Franka'' is a Franka robot equipped with the default Franka gripper, and ``Robotiq++'' is a Franka robot equipped with a custom modified Robotiq gripper with colorful padding. \textbf{Right:} Performance of fine-tuned OpenVLA and $\pi_0$ policies trained on the original Bridge and augmented Bridge data. Each policy is evaluated with 10 trials per task for each embodiment.}
    \label{fig:real_result}
\end{figure*}

\subsection{Study Findings}

Fig.~\ref{fig:sim_main_results} reports the average success rates and standard errors for each of the policies across the five tasks. Each policy is evaluated with 100 trials per task. 

\textbf{Robustness to visual perturbations.}  
Fig.~\ref{fig:sim_result_source_robot} shows performance under test-time perturbations to the source robot’s environment: (1) lighting shifts that introduce shadows, and (2) occlusions from randomly placed black rectangles. While policies trained on only the source robot degrade significantly, training with 1 and $N$ augmented robots are significantly more robust, with $N$ robot augmentation consistently achieving higher success than 1 robot augmentation. This suggests that robot augmentation enhances robustness on the original embodiment by encouraging the policy to focus on task-relevant structure (e.g., the spatial relationship between gripper and object) rather than incidental features such as arm texture or lighting cues.

\textbf{Transfer to augmented robots.}  
Among the transfer settings, training on the augmented target robot (``1× Aug (Target)'' ) achieves 26--65\% success across the four robots. Combining real and augmented data (``1× Aug (Target) + Source'') improves performance by 9--19\%. Training on all $N$ augmented robots yields the best overall performance 60\% of the time, though the gains over 1 augmented robot + source are small. 

\textbf{Generalization to unseen robots.}  
In the generalization settings, policies without robot augmentation perform poorly, as expected. While 1× robot augmentation generalizes moderately well to some robots, ($N-1$)× augmentation achieves substantially higher success across all four target robots. In many cases, ``($N-1$)× Aug (not incl. Target) + Source'' even rivals the performance of ``1× Aug (Target)'' policies trained directly on the augmented robot. This suggests that robot augmentation promotes embodiment-agnostic visual representations and spatial reasoning, and that generalization to unseen embodiments improves as we increase the diversity of robot augmentations.

Overall, these results suggest that robot augmentation improves not only pairwise transfer but also generalization and robustness. Scaling the number of augmented embodiments helps the policy learn invariances that transfer to unseen embodiments and visual conditions. These results support the view that robot augmentation is a broadly useful training strategy for generalist policies and not just a workaround for target-specific transfer.


\section{OXE-AugE: A Large Open-Source Robot Augmentation Dataset}
\label{ssec:OXE-AugE}

Motivated by the simulation results in Section~\ref{sec:sim_study}, we present \textbf{OXE-AugE}, a large-scale robot-augmented dataset derived from the Open X-Embodiment (OXE) collection \cite{open_x_embodiment_rt_x_2023}. OXE-AugE is designed to scale the benefits of robot augmentation by applying cross-painting to a broad range of tasks, scenes, and robot embodiments.

We select 16 datasets from OXE that are commonly used in training robot foundation models \cite{firoozi2025foundation, open_x_embodiment_rt_x_2023, octo_2023, kim2024openvla, black2024pi_0, bjorck2025gr00t, black2025pi05}. The original demonstrations in these datasets were each collected using a single robot---one among Franka, UR5, xArm, WidowX, Google Robot, and Jaco. We augment each dataset with up to 9 different robots: the 6 aforementioned robots, as well as Sawyer, Kinova Gen3, and KUKA. Fig.\ref{fig:cover} shows example visualizations of cross-painted augmentations, and a detailed list of the source and available augmented robots and grippers for each dataset is in the appendix.


Overall, OXE-AugE contains over 4.4M trajectories---3$\times$ larger than the original OXE dataset. It spans diverse manipulation scenes and robot-task combinations, substantially increasing embodiment diversity. The AugE-Toolkit is also open-sourced to enable the community to extend augmentation to new datasets or robots.


\begin{figure*}[t]
    \thisfloatpagestyle{empty}
    \centering
    \includegraphics[width=\textwidth,keepaspectratio]{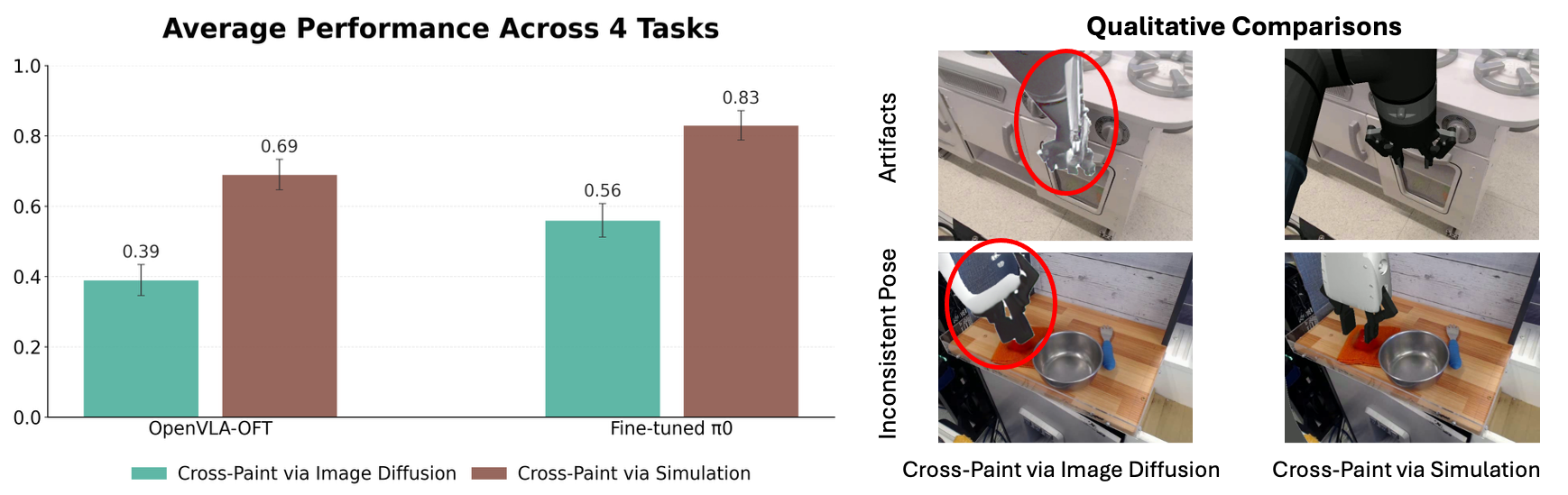}
    \caption{\textbf{Comparison with RoVi-Aug~\cite{chen2024roviaug}.} We compare our simulator-based augmentation with the RoVi-Aug diffusion pipeline. \textbf{Left:} Diffusion-based augmentation leads to a 27–30\% drop in policy success. \textbf{Right:} Qualitative examples show diffusion artifacts such as misaligned grippers and inconsistent geometry, which reduce policy performance.}
    \label{fig:real_result_diffusion_comparison}
\end{figure*}

\subsection{Physical Experiments: Fine-tuning Generalist Policies on OXE-AugE}
\label{ssec:physical_experiments}

While Section~\ref{sec:sim_study} demonstrated that robot augmentation improves transfer and robustness in the single-task setting, in this section, we evaluate whether large-scale augmentation can also benefit pretrained foundation models. 


We consider two state-of-the-art generalist policies—OpenVLA~\cite{kim2024openvla} and $\pi_0$~\cite{black2024pi_0}—and fine-tune them using OXE-AugE.  
For evaluation, we use tasks from the Bridge dataset~\cite{ebert2021bridge, walke2023bridgedata}, originally collected on a WidowX robot, and test on two embodiments:  
(1) a Franka arm with its default parallel-jaw gripper (``Franka''), which corresponds to one of the augmented robots in OXE-AugE, and  
(2) a Franka arm with a custom modified Robotiq gripper (``Robotiq++''), which features colored pads to simulate an unseen embodiment (see Fig.~\ref{fig:real_result}).  
This setup evaluates both transfer to augmented robots and generalization to unseen robot-gripper configurations.

We evaluate 4 tasks: ``Put the pot on the cloth,'' ``Put the carrot on the plate,'' ``Put the croissant into the pan,'' and ``Put the toy tiger to the left of the napkin.''  
The first three appear in Bridge, while the last is novel and absent from both Bridge and other OXE datasets, testing generalization.

For each base model, we compare ``no augmentation,'' ``2 augmented robots,'' and ``$N$ augmented robots.'' Specifically, ``2 augmented robots'' is the base model finetuned on the Franka and UR5e augmentation of OXE-AugE, and ``$N$ augmented robots'' is the base model finetuned on all augmentations. All augmented robots use their default grippers. For fair comparison, ``no augmentation'' models are lightly fine-tuned on Bridge without augmentation to match the total number of fine-tuning steps. 
For OpenVLA, we follow OpenVLA-OFT \cite{kim2025fine} and perform LoRA fine-tuning. For $\pi_0$, we use full fine-tuning works as we find it works best. Both models use $256{\times}256$ third-person observations conditioned on language instructions. More details are in the Appendix.

\subsection{Results}


Each policy is evaluated over 10 trials per task, resulting in 40 trials per embodiment. Results are summarized in Fig.~\ref{fig:real_result}.  
Both OpenVLA-OFT and $\pi_0$'s performances are relatively low when fine-tuned only on the original Bridge data, especially on Franka, due to the visual domain shift from the black WidowX gripper to the white Franka one.  
Fine-tuning on OXE-AugE significantly improves cross-embodiment performance: ``2 augmented robots'' improves success across all tasks, and $N$× augmentation yields the highest success overall.  
On average, $N$× augmentation improves performance by 24\% for OpenVLA-OFT and 45\% for $\pi_0$.  
On the novel Robotiq++ embodiment, fine-tuned policies reach 75\% (OpenVLA-OFT) and 82\% ($\pi_0$) average success, demonstrating strong generalization.

We further compare our simulator-based pipeline with the diffusion-based RoVi-Aug~\cite{chen2024roviaug}.  
Following their setup, we generate 700K paired images between WidowX and target robots using MuJoCo~\cite{todorov2012mujoco} and train diffusion models (based on Stable Diffusion~\cite{Rombach_2022_stablediffusion} and ControlNet~\cite{zhang2023adding}) to translate robot appearances.  
Fine-tuning with diffusion-generated data leads to a 27--30\% drop in final policy performance.  
As shown in Fig.~\ref{fig:real_result_diffusion_comparison}, diffusion-based augmentations sometimes produce misaligned gripper poses or geometric inconsistencies, highlighting the importance of physically accurate simulation in robot augmentation.


\section{Conclusion}
\label{sec:discussion}

In this work, we generalize robot augmentation beyond pairwise transfer and develop it into a scalable data pipeline for large-scale robot learning. By improving the cross-painting process, we make augmentation both high-quality and applicable to many existing datasets at scale. Through systematic experiments in both simulation and the real world, we demonstrate that policy performance improves with the number and diversity of augmented embodiments, yielding stronger generalization to unseen robots and greater robustness to visual perturbations. 

We introduce \textbf{OXE-AugE}, a large-scale open-source extension of the Open X-Embodiment dataset that applies a scalable and high-quality augmentation pipeline to 16 widely used datasets, expanding them to over 4.4M trajectories and 9 robot embodiments. Fine-tuning foundation models such as OpenVLA and $\pi_0$ on OXE-AugE improves success rates by up to 45\% on previously unseen robot–gripper combinations, demonstrating that explicit embodiment augmentation complements large pooled data training and promotes more robust, embodiment-agnostic reasoning.

Limitations and promising directions for future research: AugE-Toolkit performs augmentation in 2D image space using simulation replays, which is simpler and more scalable than full 3D scene reconstruction but does not model accurate object–robot occlusions. It also assumes that all augmented robots can perform the task with similar control strategies, neglecting interaction and dynamic differences across embodiments. Variations in robot size, shape, and material can lead to different feasible strategies, and our evaluation focuses mainly on pick-and-place manipulation tasks. Future work could incorporate 3D geometry and physics-aware augmentation, extend to dynamic or bimanual settings, and integrate complementary augmentations such as viewpoint, background, or object variations to enhance generalization across robots, scenes, and tasks.

\section*{Acknowledgments}\label{sec:acknowledgments}

This research was performed at the AUTOLab at UC Berkeley in affiliation with the Berkeley AI Research (BAIR) Lab. L.Y. Chen was supported by the National Science Foundation (NSF) Graduate Research Fellowship Program under Grant No. 2146752. 
The authors thank Mehdi Khfifi for some early development of our simulator, and 
Pannag Sanketi, Ted Xiao, Ashwin Balakrishna, and Quan Vuong for helpful discussions.

\newpage
\balance

\bibliographystyle{unsrtnat}  
\bibliography{references}

\newpage
\appendix
\onecolumn

\setlength{\tabcolsep}{0.6pt}
\renewcommand{\arraystretch}{0}

\clearpage

\newcommand{\chk}{\textbf{\checkmark}}
\newcommand{\markused}{\CIRCLE}

\section{Supplementary Material}
\label{sec:supplementary}

\subsection{Author contributions:} 
\textbf{G.J.} implemented the core idea, developed the simulator, labeled, generated, and curated data, trained $\pi_0$ policies, and led the physical experiments. \\
\textbf{H.P.} implemented the core idea, developed the simulator, conducted the simulation study, generated and curated data, trained OpenVLA policies, and ran some physical experiments.\\
\textbf{L.Y.C.} proposed the research, set the vision and direction, steered and aligned the team, provided guidance on all aspects of the project including the core approach, systems, and evaluations, ran some physical experiments, and wrote the paper.\\
\textbf{S.B.} developed the simulator, trained the segmentation model, generated part of the OXE-AugE data, and ran some physical experiments.\\
\textbf{Z.M.} developed the simulator, generated part of the OXE-AugE data, and contributed to paper writing.\\
\textbf{S.A.} developed the simulator used to generate part of the OXE-AugE data.\\
\textbf{C.X.} co-proposed the research with L.Y.C., advised on the project, aligned the team, provided insights and guidance on the algorithm and experiment design, generated the RoVi-Aug baseline data, and gave feedback on the paper.\\
\textbf{K.G.} supervised the project, aligned the team, and provided feedback on the approach, evaluations, and paper.\\

\subsection{OXE-AugE Dataset Details}
\label{ssec:OXE-AugEdetails}

Table~\ref{tab:dataset_robot} presents a list of the datasets in OXE-AugE. We select 16 datasets from OXE that are commonly used in training robot foundation models \cite{firoozi2025foundation, open_x_embodiment_rt_x_2023, octo_2023, kim2024openvla, black2024pi_0, bjorck2025gr00t, black2025pi05}. The original demonstrations in those datasets were collected using Franka, UR5, xArm, WidowX, Google Robot, and Jaco platforms.
A filled circle (\markused) indicates the source robot, and a check mark (\chk) indicates robots for which augmented demonstrations are available. For 14 out of the 16 datasets, all 9 robots are available. For RT-1 (Fractal) and Language Table datasets, we find most trajectories' range of the motion exceeds the workspace of WidowX, so we exclude the WidowX augmentation. Overall, OXE-AugE contains over 550{,}000 demonstrations per robot, totaling 4.4M trajectories---3$\times$ larger than the original OXE dataset.

\setlength{\fboxsep}{0pt}
\newcommand{\cell}{0.333\linewidth}
\newcommand{\HL}{FF2D55} 

\newcommand{\MaybeInc}[3]{
  \IfFileExists{images/extracted_frames/#2/#3.jpg}{%
    \includegraphics[width=#1]{images/extracted_frames/#2/#3.jpg}%
  }{%
    \fbox{\rule{0pt}{#1}\rule{#1}{0pt}}%
  }%
}


\newcommand{\CellImg}[2]{%
  \fbox{\MaybeInc{\cell}{#1}{#2}}%
}

\newcommand{\CellHL}[2]{%
  {\begingroup\color[HTML]{\HL}\fboxrule=2pt%
    \fbox{\MaybeInc{\cell}{#1}{#2}}%
  \endgroup}%
}

\newcommand{\CellMaybe}[2]{%
  \makebox[\cell][c]{\rule{0pt}{0pt}}%
}

\newcommand{\GridNine}[9]{%
  \begingroup
  \setlength{\tabcolsep}{0pt}\renewcommand{\arraystretch}{0}%
  \begin{tabular}{@{}c@{}c@{}c@{}}%
    #1 & #2 & #3\\
    #4 & #5 & #6\\
    #7 & #8 & #9
  \end{tabular}%
  \endgroup
}
\newcommand{\GridEight}[8]{%
  \begingroup
  \setlength{\tabcolsep}{0pt}\renewcommand{\arraystretch}{0}%
  \begin{tabular}{@{}c@{}c@{}c@{}}%
    #1 & #2 & #3\\
    #4 & #5 & #6\\
    \multicolumn{3}{@{}c@{}}{%
      \begin{tabular}{@{}c@{}c@{}} #7 & #8 \end{tabular}%
    }%
  \end{tabular}%
  \endgroup
}

\newcommand{\tilew}{0.24\textwidth}
\newcommand{\Tile}[2]{%
  \begin{minipage}[t]{\tilew}
    \centering
    \textbf{#1}\par\vspace{2mm}
    #2
  \end{minipage}%
}

\begin{figure}[!t]
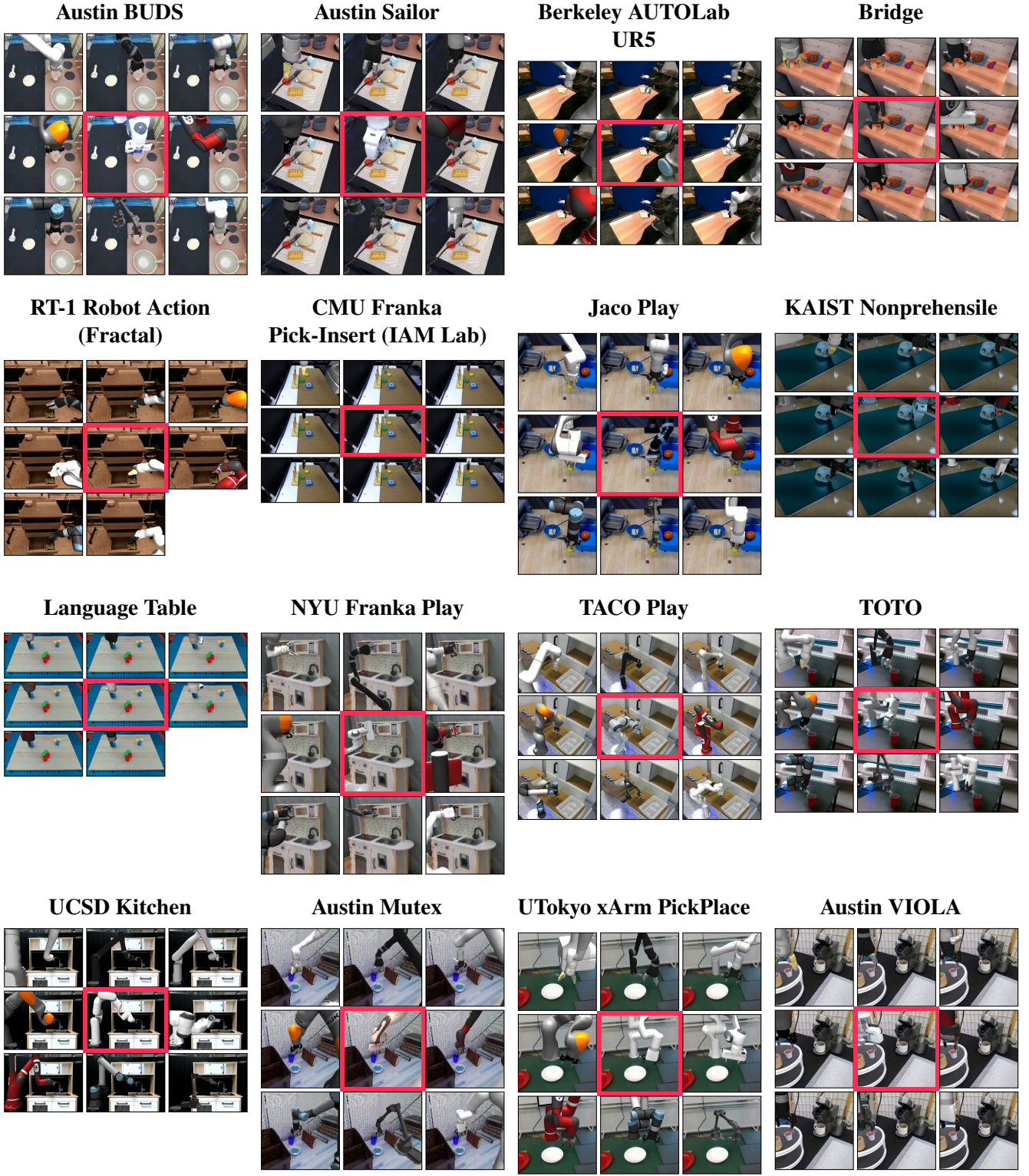

\centering
\setlength{\tabcolsep}{6pt}
\renewcommand{\arraystretch}{0}

\begin{tabular}{@{}c c c c@{}}

\Tile{Austin BUDS}{
  \GridNine{%
    \CellImg{austin_buds_dataset_converted_externally_to_rlds}{google_robot}}%
    {\CellImg{austin_buds_dataset_converted_externally_to_rlds}{jaco}}%
    {\CellImg{austin_buds_dataset_converted_externally_to_rlds}{kinova3}}%
    {\CellImg{austin_buds_dataset_converted_externally_to_rlds}{kuka_iiwa}}%
    {\CellHL{austin_buds_dataset_converted_externally_to_rlds}{panda}}%
    {\CellImg{austin_buds_dataset_converted_externally_to_rlds}{sawyer}}%
    {\CellImg{austin_buds_dataset_converted_externally_to_rlds}{ur5e}}%
    {\CellImg{austin_buds_dataset_converted_externally_to_rlds}{widowX}}%
    {\CellImg{austin_buds_dataset_converted_externally_to_rlds}{xarm7}}%
} &
\Tile{Austin Sailor}{
  \GridNine{%
    \CellImg{austin_sailor_dataset_converted_externally_to_rlds}{google_robot}}%
    {\CellImg{austin_sailor_dataset_converted_externally_to_rlds}{jaco}}%
    {\CellImg{austin_sailor_dataset_converted_externally_to_rlds}{kinova3}}%
    {\CellImg{austin_sailor_dataset_converted_externally_to_rlds}{kuka_iiwa}}%
    {\CellHL{austin_sailor_dataset_converted_externally_to_rlds}{panda}}%
    {\CellImg{austin_sailor_dataset_converted_externally_to_rlds}{sawyer}}%
    {\CellImg{austin_sailor_dataset_converted_externally_to_rlds}{ur5e}}%
    {\CellImg{austin_sailor_dataset_converted_externally_to_rlds}{widowX}}%
    {\CellImg{austin_sailor_dataset_converted_externally_to_rlds}{xarm7}}%
} &
\Tile{Berkeley AUTOLab UR5 }{
  \GridNine{%
    \CellImg{berkeley_autolab_ur5}{google_robot}}%
    {\CellImg{berkeley_autolab_ur5}{jaco}}%
    {\CellImg{berkeley_autolab_ur5}{kinova3}}%
    {\CellImg{berkeley_autolab_ur5}{kuka_iiwa}}%
    {\CellHL{berkeley_autolab_ur5}{ur5e}}%
    {\CellImg{berkeley_autolab_ur5}{panda}}%
    {\CellImg{berkeley_autolab_ur5}{sawyer}}%
    {\CellImg{berkeley_autolab_ur5}{widowX}}%
    {\CellImg{berkeley_autolab_ur5}{xarm7}}%
} &
\Tile{Bridge}{
  \GridNine{%
    \CellImg{bridge}{google_robot}}%
    {\CellImg{bridge}{jaco}}%
    {\CellImg{bridge}{kinova3}}%
    {\CellImg{bridge}{kuka_iiwa}}%
    {\CellHL{bridge}{widowX}}%
    {\CellImg{bridge}{panda}}%
    {\CellImg{bridge}{sawyer}}%
    {\CellImg{bridge}{ur5e}}%
    {\CellImg{bridge}{xarm7}}%
}
\\[6pt]
\multicolumn{4}{@{}c@{}}{\rule{0pt}{12pt}}\\
\Tile{RT-1 Robot Action (Fractal)}{
  \GridNine{%
    \CellImg{fractal20220817_data}{jaco}}%
    {\CellImg{fractal20220817_data}{kinova3}}%
    {\CellImg{fractal20220817_data}{kuka_iiwa}}%
    {\CellImg{fractal20220817_data}{panda}}%
    {\CellHL{fractal20220817_data}{google_robot}}%
    {\CellImg{fractal20220817_data}{sawyer}}%
    {\CellImg{fractal20220817_data}{ur5e}}%
    {\CellImg{fractal20220817_data}{xarm7}}%
    {\CellMaybe{fractal20220817_data}{widowX}}%
} &
\Tile{CMU Franka Pick-Insert (IAM Lab)}{
  \GridNine{%
    \CellImg{iamlab_cmu_pickup_insert_converted_externally_to_rlds}{google_robot}}%
    {\CellImg{iamlab_cmu_pickup_insert_converted_externally_to_rlds}{jaco}}%
    {\CellImg{iamlab_cmu_pickup_insert_converted_externally_to_rlds}{kinova3}}%
    {\CellImg{iamlab_cmu_pickup_insert_converted_externally_to_rlds}{kuka_iiwa}}%
    {\CellHL{iamlab_cmu_pickup_insert_converted_externally_to_rlds}{panda}}%
    {\CellImg{iamlab_cmu_pickup_insert_converted_externally_to_rlds}{sawyer}}%
    {\CellImg{iamlab_cmu_pickup_insert_converted_externally_to_rlds}{ur5e}}%
    {\CellImg{iamlab_cmu_pickup_insert_converted_externally_to_rlds}{widowX}}%
    {\CellImg{iamlab_cmu_pickup_insert_converted_externally_to_rlds}{xarm7}}%
} &
\Tile{Jaco Play}{
  \GridNine{%
    \CellImg{jaco_play}{google_robot}}%
    {\CellImg{jaco_play}{kinova3}}%
    {\CellImg{jaco_play}{kuka_iiwa}}%
    {\CellImg{jaco_play}{panda}}%
    {\CellHL{jaco_play}{jaco}}%
    {\CellImg{jaco_play}{sawyer}}%
    {\CellImg{jaco_play}{ur5e}}%
    {\CellImg{jaco_play}{widowX}}%
    {\CellImg{jaco_play}{xarm7}}%
} &
\Tile{KAIST Nonprehensile}{
  \GridNine{%
    \CellImg{kaist_nonprehensile_converted_externally_to_rlds}{google_robot}}%
    {\CellImg{kaist_nonprehensile_converted_externally_to_rlds}{jaco}}%
    {\CellImg{kaist_nonprehensile_converted_externally_to_rlds}{kinova3}}%
    {\CellImg{kaist_nonprehensile_converted_externally_to_rlds}{kuka_iiwa}}%
    {\CellHL{kaist_nonprehensile_converted_externally_to_rlds}{panda}}%
    {\CellImg{kaist_nonprehensile_converted_externally_to_rlds}{sawyer}}%
    {\CellImg{kaist_nonprehensile_converted_externally_to_rlds}{ur5e}}%
    {\CellImg{kaist_nonprehensile_converted_externally_to_rlds}{widowX}}%
    {\CellImg{kaist_nonprehensile_converted_externally_to_rlds}{xarm7}}%
}
\\[6pt]
\multicolumn{4}{@{}c@{}}{\rule{0pt}{12pt}}\\
\Tile{Language Table}{
  \GridNine{%
    \CellImg{language_table}{google_robot}}%
    {\CellImg{language_table}{jaco}}%
    {\CellImg{language_table}{kinova3}}%
    {\CellImg{language_table}{kuka_iiwa}}%
    {\CellHL{language_table}{xarm7}}%
    {\CellImg{language_table}{panda}}%
    {\CellImg{language_table}{sawyer}}%
    {\CellImg{language_table}{ur5e}}%
    {\CellMaybe{language_table}{widowX}}%
} &
\Tile{NYU Franka Play}{
  \GridNine{%
    \CellImg{nyu_franka_play_dataset_converted_externally_to_rlds}{google_robot}}%
    {\CellImg{nyu_franka_play_dataset_converted_externally_to_rlds}{jaco}}%
    {\CellImg{nyu_franka_play_dataset_converted_externally_to_rlds}{kinova3}}%
    {\CellImg{nyu_franka_play_dataset_converted_externally_to_rlds}{kuka_iiwa}}%
    {\CellHL{nyu_franka_play_dataset_converted_externally_to_rlds}{panda}}%
    {\CellImg{nyu_franka_play_dataset_converted_externally_to_rlds}{sawyer}}%
    {\CellImg{nyu_franka_play_dataset_converted_externally_to_rlds}{ur5e}}%
    {\CellImg{nyu_franka_play_dataset_converted_externally_to_rlds}{widowX}}%
    {\CellImg{nyu_franka_play_dataset_converted_externally_to_rlds}{xarm7}}%
} &
\Tile{TACO Play}{
  \GridNine{%
    \CellImg{taco_play}{google_robot}}%
    {\CellImg{taco_play}{jaco}}%
    {\CellImg{taco_play}{kinova3}}%
    {\CellImg{taco_play}{kuka_iiwa}}%
    {\CellHL{taco_play}{panda}}%
    {\CellImg{taco_play}{sawyer}}%
    {\CellImg{taco_play}{ur5e}}%
    {\CellImg{taco_play}{widowX}}%
    {\CellImg{taco_play}{xarm7}}%
} &
\Tile{TOTO}{
  \GridNine{%
    \CellImg{toto}{google_robot}}%
    {\CellImg{toto}{jaco}}%
    {\CellImg{toto}{kinova3}}%
    {\CellImg{toto}{kuka_iiwa}}%
    {\CellHL{toto}{panda}}%
    {\CellImg{toto}{sawyer}}%
    {\CellImg{toto}{ur5e}}%
    {\CellImg{toto}{widowX}}%
    {\CellImg{toto}{xarm7}}%
}
\\[6pt]
\multicolumn{4}{@{}c@{}}{\rule{0pt}{12pt}}\\
\Tile{UCSD Kitchen}{
  \GridNine{%
    \CellImg{ucsd_kitchen_dataset_converted_externally_to_rlds}{google_robot}}%
    {\CellImg{ucsd_kitchen_dataset_converted_externally_to_rlds}{jaco}}%
    {\CellImg{ucsd_kitchen_dataset_converted_externally_to_rlds}{kinova3}}%
    {\CellImg{ucsd_kitchen_dataset_converted_externally_to_rlds}{kuka_iiwa}}%
    {\CellHL{ucsd_kitchen_dataset_converted_externally_to_rlds}{xarm7}}%
    {\CellImg{ucsd_kitchen_dataset_converted_externally_to_rlds}{panda}}%
    {\CellImg{ucsd_kitchen_dataset_converted_externally_to_rlds}{sawyer}}%
    {\CellImg{ucsd_kitchen_dataset_converted_externally_to_rlds}{ur5e}}%
    {\CellImg{ucsd_kitchen_dataset_converted_externally_to_rlds}{widowX}}%
} &
\Tile{Austin Mutex}{
  \GridNine{%
    \CellImg{utaustin_mutex}{google_robot}}%
    {\CellImg{utaustin_mutex}{jaco}}%
    {\CellImg{utaustin_mutex}{kinova3}}%
    {\CellImg{utaustin_mutex}{kuka_iiwa}}%
    {\CellHL{utaustin_mutex}{panda}}%
    {\CellImg{utaustin_mutex}{sawyer}}%
    {\CellImg{utaustin_mutex}{ur5e}}%
    {\CellImg{utaustin_mutex}{widowX}}%
    {\CellImg{utaustin_mutex}{xarm7}}%
} &
\Tile{UTokyo xArm PickPlace}{
  \GridNine{%
    \CellImg{utokyo_xarm_pick_and_place_converted_externally_to_rlds}{google_robot}}%
    {\CellImg{utokyo_xarm_pick_and_place_converted_externally_to_rlds}{jaco}}%
    {\CellImg{utokyo_xarm_pick_and_place_converted_externally_to_rlds}{kinova3}}%
    {\CellImg{utokyo_xarm_pick_and_place_converted_externally_to_rlds}{kuka_iiwa}}%
    {\CellHL{utokyo_xarm_pick_and_place_converted_externally_to_rlds}{xarm7}}%
    {\CellImg{utokyo_xarm_pick_and_place_converted_externally_to_rlds}{panda}}%
    {\CellImg{utokyo_xarm_pick_and_place_converted_externally_to_rlds}{sawyer}}%
    {\CellImg{utokyo_xarm_pick_and_place_converted_externally_to_rlds}{ur5e}}%
    {\CellImg{utokyo_xarm_pick_and_place_converted_externally_to_rlds}{widowX}}%
} &
\Tile{Austin VIOLA}{
  \GridNine{%
    \CellImg{viola}{google_robot}}%
    {\CellImg{viola}{jaco}}%
    {\CellImg{viola}{kinova3}}%
    {\CellImg{viola}{kuka_iiwa}}%
    {\CellHL{viola}{panda}}%
    {\CellImg{viola}{sawyer}}%
    {\CellImg{viola}{ur5e}}%
    {\CellImg{viola}{widowX}}%
    {\CellImg{viola}{xarm7}}%
}
\end{tabular}
\caption{Example images in OXE-AugE. \footnotesize
  {\begingroup\setlength{\fboxsep}{0pt}\fboxrule=2pt\color[HTML]{\HL}\fbox{\rule{8pt}{8pt}}\endgroup}
  \;=\; Source robot (center cell, highlighted)
}
\label{fig:oxe_aug_example_visualization}
\end{figure}

\begin{table*}[h]
    \centering
    \small
    \setlength{\tabcolsep}{6pt}
    \renewcommand{\arraystretch}{1.15}
    \begin{tabularx}{\textwidth}{@{}l*{9}{>{\centering\arraybackslash}X}@{}}
        \toprule
        \textbf{Dataset} & \textbf{Panda} & \textbf{UR5e} & \textbf{Xarm7} & \textbf{Google} & \textbf{WidowX} & \textbf{Sawyer} & \textbf{Kinova3} & \textbf{IIWA} & \textbf{Jaco} \\
        \midrule
        Berkeley AUTOLab UR5~\cite{BerkeleyUR5Website} & \chk & \markused & \chk & \chk & \chk & \chk & \chk & \chk & \chk \\
        TACO Play~\cite{zhou2023train} & \markused & \chk & \chk & \chk & \chk & \chk & \chk & \chk & \chk \\
        Austin BUDS~\cite{zhu2022bottom} & \markused & \chk & \chk & \chk & \chk & \chk & \chk & \chk & \chk \\
        Austin Mutex~\cite{shah2023mutex} & \markused & \chk & \chk & \chk & \chk & \chk & \chk & \chk & \chk \\
        Austin Sailor~\cite{nasiriany2022learning} & \markused & \chk & \chk & \chk & \chk & \chk & \chk & \chk & \chk \\
        CMU Franka Pick-Insert~\cite{saxena2023multiresolution} & \markused & \chk & \chk & \chk & \chk & \chk & \chk & \chk & \chk \\
        KAIST Nonprehensile~\cite{kimpre} & \markused & \chk & \chk & \chk & \chk & \chk & \chk & \chk & \chk \\
        NYU Franka Play~\cite{cui2022play} & \markused & \chk & \chk & \chk & \chk & \chk & \chk & \chk & \chk \\
        TOTO~\cite{zhou2023train} & \markused & \chk & \chk & \chk & \chk & \chk & \chk & \chk & \chk \\
        UTokyo xArm PickPlace~\cite{matsushima2023weblab} & \chk & \chk & \markused & \chk & \chk & \chk & \chk & \chk & \chk \\
        UCSD Kitchen~\cite{ucsd_kitchens} & \chk & \chk & \markused & \chk & \chk & \chk & \chk & \chk & \chk \\
        Austin VIOLA~\cite{zhu2022viola} & \markused & \chk & \chk & \chk & \chk & \chk & \chk & \chk & \chk \\
        Bridge~\cite{walke2023bridgedata} & \chk & \chk & \chk & \chk & \markused & \chk & \chk & \chk & \chk \\
        RT-1 Robot Action~\cite{brohan2023rt1} & \chk & \chk & \chk & \markused &  & \chk & \chk & \chk & \chk \\
        Jaco Play~\cite{dass2023jacoplay} & \chk & \chk & \chk & \chk & \chk & \chk & \chk & \chk & \markused \\
        Language Table~\cite{lynch2023interactive} & \chk & \chk & \markused & \chk &  & \chk & \chk & \chk & \chk \\
        \bottomrule
    \end{tabularx}
    \caption{\textbf{Source and augmented robots in the OXE-AugE dataset.}
    A filled circle (\markused) indicates the \textbf{source robot}, and a check mark (\chk) indicates robots for which augmented demonstrations are available.}
    \label{tab:dataset_robot}
\end{table*}

In both the original OXE dataset and OXE-AugE, the Franka robot uses the default Franka Hand, UR5e uses the Robotiq 2f-85 gripper, xArm7 uses UFACTORY xArm Gripper G2, Google Robot uses the custom 2-finger gripper, Jaco uses the Kinova KG-3 gripper, and WidowX 250 uses the default parallel-jaw gripper. Additionally, OXE-AugE uses the Robotiq 2f-85 gripper for KUKA iiwa and Kinova3, and the Rethink Gripper for the Sawyer robot. We use these grippers as they are the most common types used for each robot, but switching grippers is also easy to do.

\begin{figure*}[h!]
    \thisfloatpagestyle{empty}
    \centering
    \includegraphics[width=\textwidth,keepaspectratio]{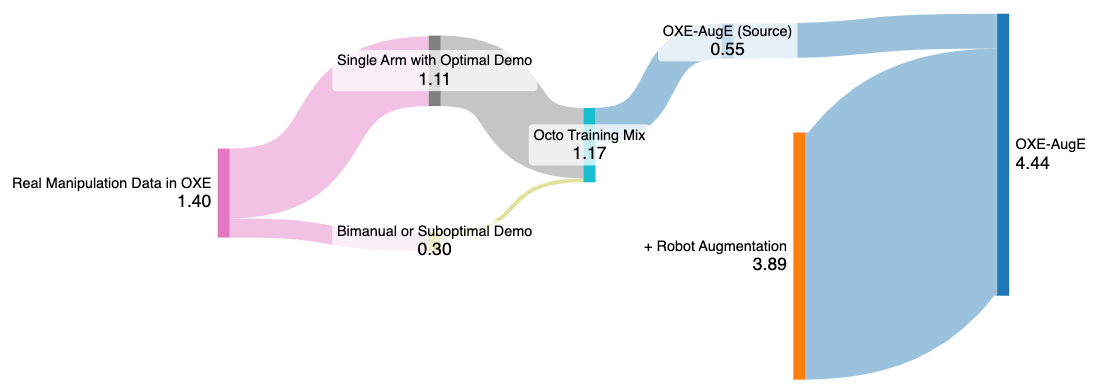}
    \caption{\textbf{Sankey diagram illustrating the data sources of OXE-AugE (numbers in millions of trajectories).} OXE v1.1 contains about 1.4M real robot manipulation trajectories. Starting from there, we select 14 datasets out of the 25 datasets in the Octo Training Mix, collectively counting for 58\% of the total weights. We also include 2 other high-quality datasets (UTokyo xArm PickPlace and KAIST Nonprehensile Objects) that are not used in Octo. This set of 0.55M trajectories forms the source datasets for OXE-AugE. After robot augmentation, the total OXE-AugE consists of 4.44M trajectories.}
    \label{fig:sankey_diagram}
\end{figure*}

Figure~\ref{fig:oxe_aug_example_visualization} shows example images in OXE-AugE. Figure~\ref{fig:sankey_diagram} illustrates the data sources of OXE-AugE and its relationship to OXE and the Octo training mixture. OXE v1.0 contains about 1.4M trajectories. V1.1 expands the total number of trajectories to 2.4M, however, only 1.4M of them are real robot data with manipulation or mobile manipulation (the remaining are sim, navigation, locomotion, human, or VQA data). Starting from there, we select 14 datasets out of the 25 datasets in the Octo Training Mix, collectively counting for 58\% of the total weights. We also include 2 other high-quality datasets (UTokyo xArm PickPlace and KAIST Nonprehensile Objects) that are not used in Octo. Together, these 16 datasets include 0.55M trajectories and form the source of OXE-AugE. After augmenting each dataset with 8 or 9 robots, the total OXE-AugE consists of 4.44M trajectories.

\begin{figure*}[h!]
    \centering
    \includegraphics[width=\textwidth,keepaspectratio]{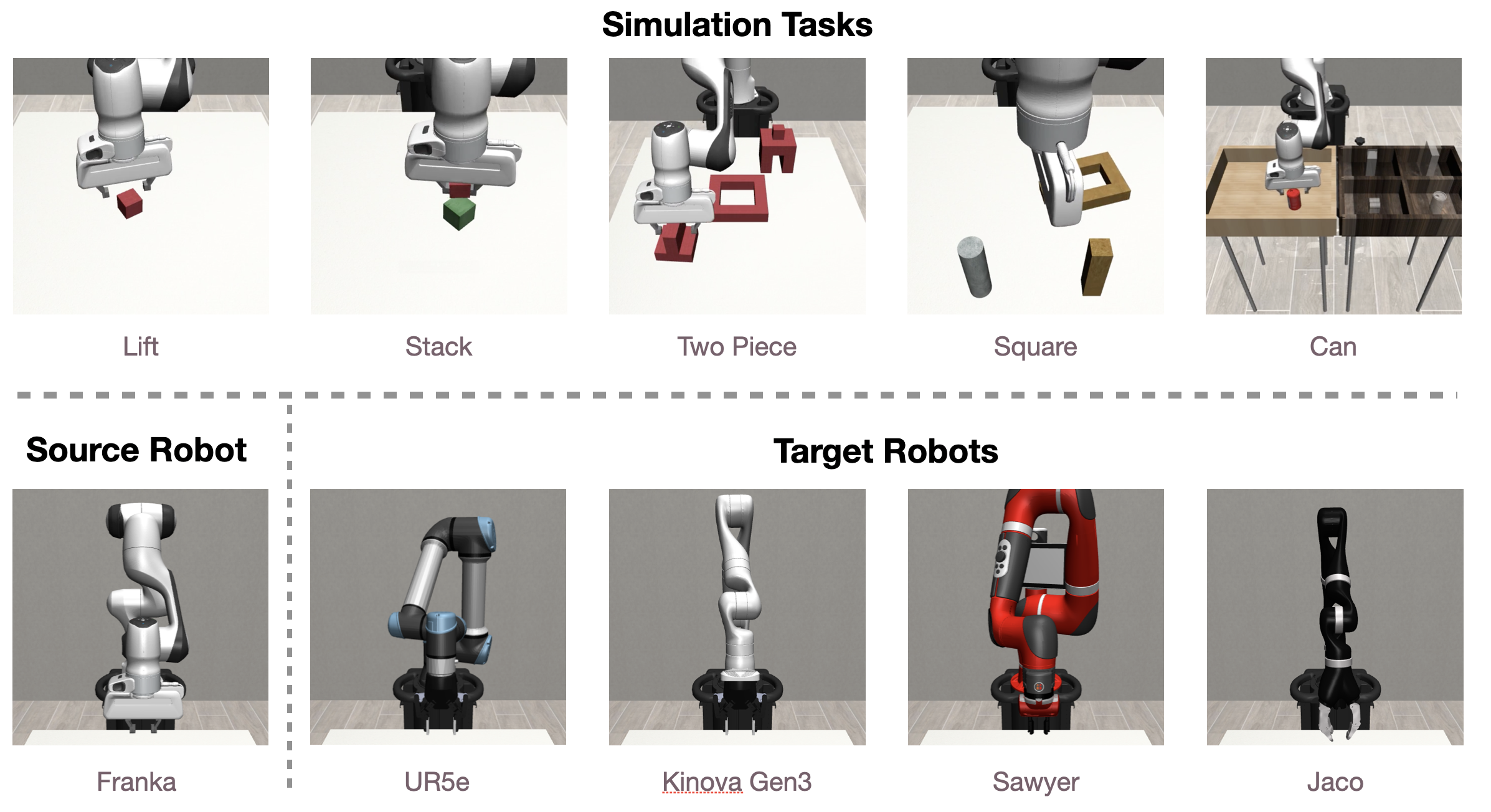}
    \caption{\textbf{Simulation Tasks and Robots.} We use the Robosuite environment with 5 tasks: Lift, Stack, Two
Piece Assembly, Square Peg Insertion tasks, and Can Pick-and-Place. The demonstration data is performed on a Franka robot, and we evaluate on 4 target robots: UR5e, Kinova Gen3, Sawyer, and Jaco. Jaco has a 3-jaw gripper while the other robots have parallel-jaw grippers.}
    \label{fig:sim_robot_tasks}
\end{figure*}

\subsection{Simulation Tasks and Robots Visualization}
\label{ssec:sim_tasks_robots}

Fig.~\ref{fig:sim_robot_tasks} illustrates the simulation tasks and robots. We use the Robosuite environment with 5 tasks: Lift, Stack, Two
Piece Assembly, Square Peg Insertion tasks, and Can Pick-and-Place. The demonstration data is performed on a Franka robot, and we evaluate on 4 target robots: UR5e, Kinova Gen3, Sawyer, and Jaco. Jaco has a 3-jaw gripper while the other robots have 2-jaw grippers.

\subsection{Policy Training Details}
\label{ssec:policy_training}

For simulation experiments, we train diffusion policies \cite{chi2023diffusion} using RoboMimic~\cite{robomimic2021}. Each policy is trained on 200 demonstrations for a single task from scratch. The policy architecture consists of a non-pretrained ResNet18 \cite{he2016deep} vision encoder and a 1D convolutional neural network (CNN) action denoiser, connected through FiLM \cite{perez2017film}. All policies are trained with a learning rate of 1e-4, batch size of 16, and for 250k steps. The visual inputs are 84x84, with random crop data augmentation during training.

For physical experiments, we fine-tune OpenVLA and $\pi_0$. For OpenVLA, we follow OpenVLA-OFT \cite{kim2025fine} and perform LoRA fine-tuning \cite{hu2022lora} with a learning rate of 5e-4, batch size of 8, for 25k steps. For $\pi_0$, we perform full parameter fine-tuning with a learning rate of 5e-5, batch size of 32, for 20k steps. Both models take in a third-person observation of 256$\times$256 resolution, and are conditioned on the language instructions.

\subsection{Physical Experiment Details}
\label{ssec:experiments_details}

In physical experiments, we use a Franka FR3 robot and a ZED 2 camera. We place the camera in roughly the same pose as that in the Bridge dataset, and crop the images to match the field of view of the Logitech camera used in Bridge. To address the challenges of different controller dynamics between the robots, we follow OpenVLA and train on the delta states of the WidowX robot instead of the action targets. During inference, we let the Franka controller reach the desired target state before executing the next action. To prevent compounding error, the policy takes in the predicted state instead of the actual state as inputs \cite{tang2023industreal}.

For the 4 tasks, we use the following metrics for measuring the policy performance: we give a score of 0.3 if the robot moves towards the right direction the object and attempts a grasp, a score of 0.5 if the robot successfully grasps the object, a score of 0.8 if the robot carries the object and moves towards the correct destination location, and a score of 1 if the robot successfully place the object in the target position. We perform 10 trials per task, resulting in 40 trials per policy for each robot embodiment. We compute the mean and standard error of the scores for each of the 6 policies on each target embodiment in Fig.~\ref{fig:real_result}.


\end{document}